\documentclass[a4paper, 10pt, conference]{ieeeconf}      

\IEEEoverridecommandlockouts                              

\pdfoutput=1

\usepackage{amssymb}  
\usepackage{subfigure}
\usepackage{caption}
\usepackage{amsmath} 
\usepackage{orcidlink}

\title{ \LARGE \bf Panoptic-SLAM: Visual SLAM in Dynamic Environments using Panoptic Segmentation}

\author{ Gabriel Fischer Abati\orcidlink{0000-0001-8158-3065}$^{1,2*}$, João Carlos Virgolino Soares\orcidlink{0000-0002-6278-378X}$^{1*}$, Vivian Suzano Medeiros\orcidlink{0000-0002-3866-8538}$^{1,3}$, \\ Marco Antonio Meggiolaro\orcidlink{0000-0002-6240-8189}$^{2}$ and Claudio Semini\orcidlink{0000-0002-3034-4686}$^{1}$
\thanks{*The authors contributed equally to this work}
\thanks{$^{1}$ Dynamic Legged Systems Lab, Istituto Italiano di Tecnologia, Italy
        {\tt\small gabriel.fischer@iit.it} /
        {\tt\small joao.virgolino@iit.it} /
        {\tt\small claudio.semini@iit.it}}%
\thanks{$^{2}$ Department of Mechanical Engineering at the Pontifical Catholic University of Rio de Janeiro, Brazil  
{\tt\small meggi@puc-rio.br}}
\thanks{$^{3}$ Department of Mechanical Engineering, University of S\~ao Paulo, Brazil {\tt\small viviansuzano@usp.br}}}

\begin{document}

\maketitle

\begin{abstract}

The majority of visual SLAM systems are not robust in dynamic scenarios. The ones that deal with dynamic objects in the scenes usually rely on deep-learning-based methods to detect and filter these objects. However, these methods cannot deal with unknown moving objects. This work presents Panoptic-SLAM, an open-source visual SLAM system robust to dynamic environments, even in the presence of unknown objects. It uses panoptic segmentation to filter dynamic objects from the scene during the state estimation process. Panoptic-SLAM is based on ORB-SLAM3, a state-of-the-art SLAM system for static environments. The implementation was tested using real-world datasets and compared with several state-of-the-art systems from the literature, including DynaSLAM, DS-SLAM, SaD-SLAM, PVO and FusingPanoptic. For example, Panoptic-SLAM is on average four times more accurate than PVO, the most recent panoptic-based approach for visual SLAM. Also, experiments were performed using a quadruped robot with an RGB-D camera to test the applicability of our method in real-world scenarios. The tests were validated by a ground-truth created with a motion capture system.

\end{abstract}

\section{INTRODUCTION}

Simultaneous Localization and Mapping (SLAM) is a crucial part of the software frameworks used in many autonomous vehicles, such as drones and mobile robots, as it enables them to navigate and operate in previously unknown or unstructured environments. It consists of constructing a map of an unknown environment while simultaneously determining the position of the robot within that environment.

Visual SLAM systems are receiving increasing attention in the robotics community due to the low cost and richness of information provided by cameras. There are several visual SLAM systems in the literature with high precision using monocular (e.g. ORBSLAM~\cite{orbslam}, LSD-SLAM~\cite{LSD-SLAM}), stereo (e.g. ORBSLAM2~\cite{orbslam2}) and RGB-D cameras (e.g. RGBDSLAM~\cite{endres}). 

However, these SLAM systems are not prepared to work in scenarios with moving/dynamic objects, which results in inaccurate localization and inconsistent mapping. To address this, there are several approaches to incorporate dynamic elements into visual SLAM. Recently, deep learning-based methods have been explored for SLAM in dynamic environments for providing high-level information about the scenes.

\begin{figure}[t]
  \subfigure[  \hspace{0.05cm}    Initial feature detection]{%
    \includegraphics[width=.473\linewidth]{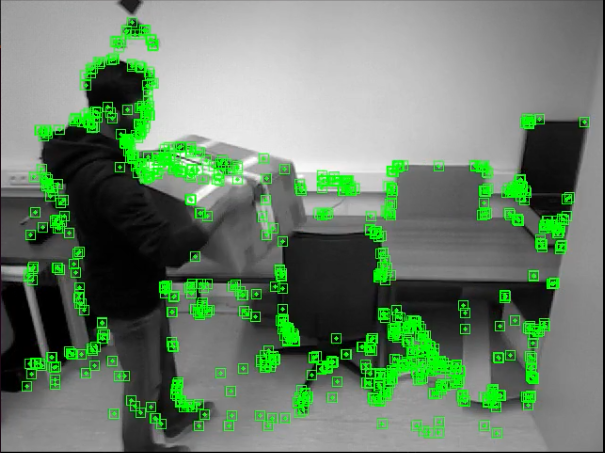}
    \label{fig:1} 
  } 
  \subfigure[ \hspace{0.05cm}   Panoptic Segmentation]{%
    \includegraphics[width=.473\linewidth]{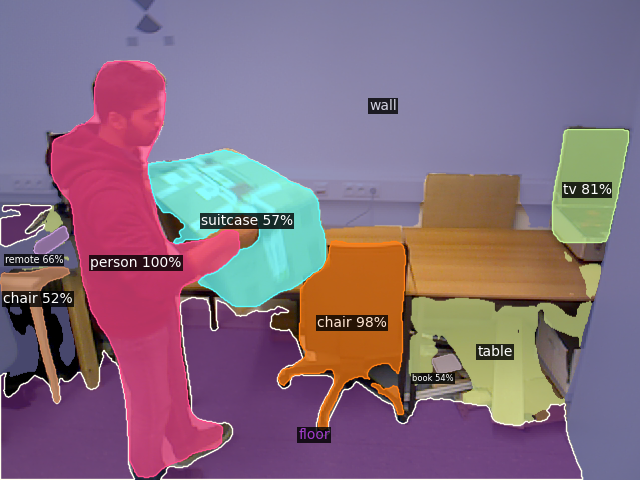}
    \label{fig:2} 
  } 

  \subfigure[  \hspace{0.05cm}    Moving Masks]{%
    \includegraphics[width=.473\linewidth]{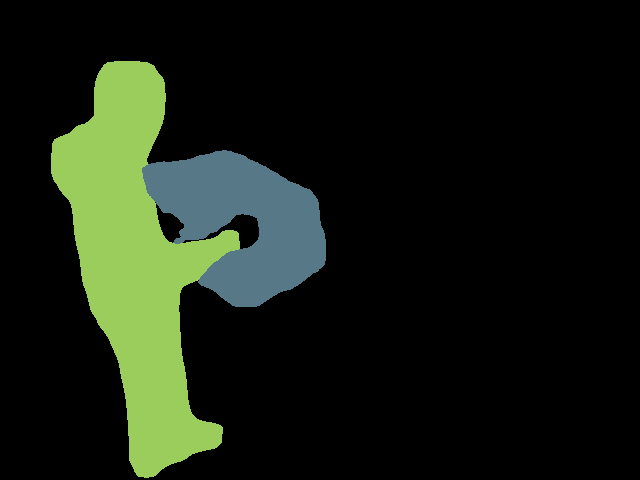}
    \label{fig:3} 
  } 
  \subfigure[ \hspace{0.05cm}   Filtering]{%
    \includegraphics[width=.473\linewidth]{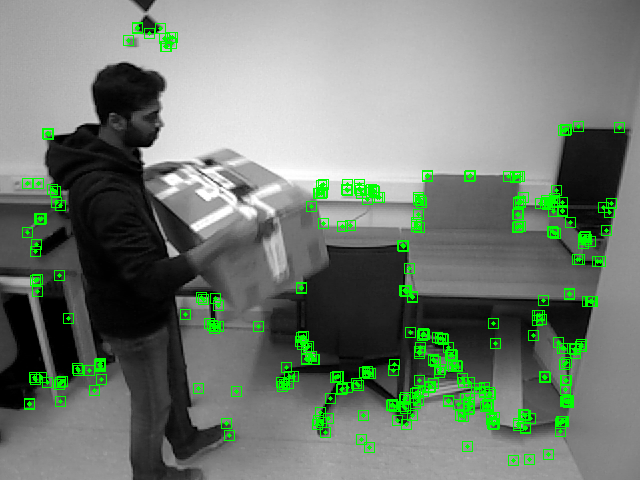}
    \label{fig:4} 
  } 
  \caption{Overview of the Panoptic-SLAM for filtering dynamic objects. In (a) initial feature detection, (b) shows the output of panoptic segmentation, (c) shows the masks of moving instances (person and box), and (d) shows the keypoints belonging to moving objects successfully filtered, even though there is no cardboard box in the list of trained classes } 
  \label{fig:firstpage}
\end{figure}

Some methods, e.g. SGC-VSLAM~\cite{sgc}, use object detection, such as YOLO~\cite{yolov3} to detect and track labeled objects to then filter out the keypoints of the dynamic objects. DS-SLAM \cite{dsslam} and DCS-SLAM \cite{dcs-slam} use semantic segmentation combined with epipolar geometry to filter dynamic keypoints. Other methods, e.g. DynaSLAM~\cite{dynaslam}, SAD-SLAM~\cite{sadslam} and DOT-MASK \cite{dotmask} rely on instance segmentation techniques, such as Mask R-CNN~\cite{mask} or YOLACT~\cite{yolact}. The main problem with all previous approaches is the necessity to have a pre-determined number of object classes that can be detected, and consequently, filtered. In other words, if a non-labeled object is moving in the scene, it would not be filtered and its features would cause a drift in the localization and outliers in the map. 

Panoptic segmentation \cite{panoptic} is a computer vision task that combines both instance segmentation and semantic segmentation. In instance segmentation, objects in the image are identified individually and segmented at a pixel level, while in semantic segmentation, each pixel in the image is labeled with a semantic category. The goal of panoptic segmentation is to unify these two tasks into a single one, where all pixels in the image are labeled with either an instance label, indicating which object it belongs to, or a semantic label, indicating which category it belongs to.

Recently some visual SLAM methods have been proposed using panoptic segmentation to handle dynamic scenarios. One example is PVO~\cite{PVO}, which uses the visual odometry~(VO) system from DROID-SLAM \cite{DROIDSLAM}, and fuses panoptic segmentation with optical flow and depth measurements to create a panoptic-aware optical flow representation of the environment. Although its VO module can obtain accurate camera poses, the authors did not specify how the unknown objects from the panoptic segmentation model affects their system. Moreover, PVO can perform robustly in dynamic scenes, but it ignores loop closure detection when the camera returns to a previously visited location \cite{PVO}.

In this work, we present Panoptic-SLAM\footnote[1]{\url{https://github.com/iit-DLSLab/Panoptic-SLAM}}, a visual SLAM for dynamic environments that uses panoptic segmentation to detect and filter moving objects, as shown in Fig. \ref{fig:firstpage}. The contributions of the paper can be summarized as follows.

\begin{itemize}
    \item A novel visual SLAM system based on panoptic segmentation that can perform robust localization in dynamic environments in the presence of unknown and unlabeled moving objects. The system can work with either monocular, stereo and RGB-D cameras since it only needs RGB image information.

    \item Tests using the Bonn RGB-D dynamic \cite{re-fusion} dataset, which explicitly contains unknown dynamic objects in the scene, and with the TUM RGB-D dataset \cite{benchmark}, obtaining the best results in several sequences compared to state-of-the-art methods, such as DynaSLAM \cite{dynaslam}, PVO \cite{PVO} and FusingPanoptic \cite{fusingpanoptic}.

    \item Experimental tests performed with an RGB-D camera attached to a quadruped robot moving in an indoor environment in a scenario containing moving and static unknown objects, such as other quadruped robots. A Vicon system was used to get the ground truth for accuracy evaluation. The new dataset was used to test the proposed approach, which proved to be on average 2 times more accurate compared to DynaSLAM. 
\end{itemize}

\section{RELATED WORK}

 Dealing with dynamic objects is a significant challenge in visual SLAM research. Most researchers treat dynamic objects as outliers, and several visual SLAM systems have been presented to address these outliers. These solutions can be broadly categorized into two main groups: geometry-based and learning-based methods. Geometry-based methods rely on classic computer vision techniques to detect and filter dynamic content. In general, they have inferior accuracy compared to learning-based methods. Their main advantage is to not require prior knowledge about the objects in the scene.
 
 Learning-based methods, on the other hand, require a pre-trained model. They usually use object detection, semantic segmentation, instance segmentation, or, more recently, panoptic segmentation. Li et al. \cite{haili2022} introduce a system that combines object detection and a geometry-based approach. The system is also based on ORB-SLAM2 with DeepSort tracking and epipolar geometry to detect static points of each object in the scene.

DS-SLAM \cite{dsslam} combines the SegNet semantic segmentation network with a moving consistency check algorithm based on optical flow to reduce the impact of dynamic objects, and generates a dense semantic octo-tree map from the environment.

Changing-SLAM \cite{changing-slam} uses a YOLOv4 framework trained with the COCO dataset to detect objects in the scene and employs an extended Kalman filter to estimate their velocity, removing the keypoints belonging to moving objects. Although robust to both dynamic and changing environments, Changing-SLAM uses a feature repopulation method coupled with object detection instead of a segmentation network, and it does not present any specific mechanism to handle unknown objects.

None of the previously cited methods can deal with unknown moving objects. Ji et al.~\cite{ji2021} presented a Semantic RGB-D SLAM approach that operates in dynamic environments by extracting semantic information solely from keyframes. Despite being able to deal with unknown objects using k-means and depth reprojection, their accuracy is lower than other methods, such as DynaSLAM, in environments with people. DynaSLAM \cite{dynaslam} is one of the first works to use an instance segmentation model \cite{mask} to detect people in the scene at a pixel-level, which it then uses to filter dynamic features. It is one of the visual SLAM systems with the best accuracy in the TUM RGB-D dataset \cite{benchmark}.

Yuan and Chen proposed SaD-SLAM \cite{sadslam}, which integrates depth information and Mask R-CNN instance segmentation to identify dynamic features in images. The algorithm classifies each feature point as dynamic, static, or static and movable. SaD-SLAM exhibits high accuracy, surpassing DynaSLAM in certain scenarios. However, its main limitation is that the semantic segmentation is conducted offline. Vincent et al. presented DOT-Mask~\cite{dotmask}, which uses an extended Kalman filter to track dynamic objects in the scene, identified by an instance segmentation module. The authors aimed to develop a faster SLAM system at the expense of lower accuracy when compared to other approaches. The main problem of instance segmentation-based approaches for visual pose estimation is the lack of background information, which decreases the possibility to infer about the existence of unknown moving objects. The use of keypoints belonging to an unknown moving object for tracking might cause failures in pose estimation.

Recently, there has been an increasing use of panoptic segmentation in many applications. Semantic scene understanding with Lidar sensors \cite{Besic_2022} and 3D mapping \cite{miao2023volumetric} are examples of the usage of the panoptic segmentation outside the scope of SLAM. Panoptic segmentation is also used in visual state estimation systems, such as the previously mentioned PVO \cite{PVO}. Analogous to PVO, SVG-LOOP \cite{Yuan2021} presents a loop-closure detection algorithm that uses a combination of a semantic bag-of-words model processed with panoptic segmentation to minimize the impact of dynamic features, and a semantic landmark vector model to encode the geometric connections within the semantic graph.  

Zhu et al. developed FusingPanoptic \cite{fusingpanoptic}, a method based on ORB-SLAM2 that incorporates panoptic segmentation and geometry information. To minimize the impact of unknown moving objects, they propose a dynamic object classification approach based on epipolar geometry. Despite claiming robustness against unknown moving objects, this was not numerically evaluated in their paper. Also, they eliminate \textit{a priori} all keypoints belonging to what they consider highly dynamic, without considering known movable objects that can be static (vehicles, for instance).

\section{METHODOLOGY}

\begin{figure*}
	\centering
        \includegraphics[scale=0.3825]{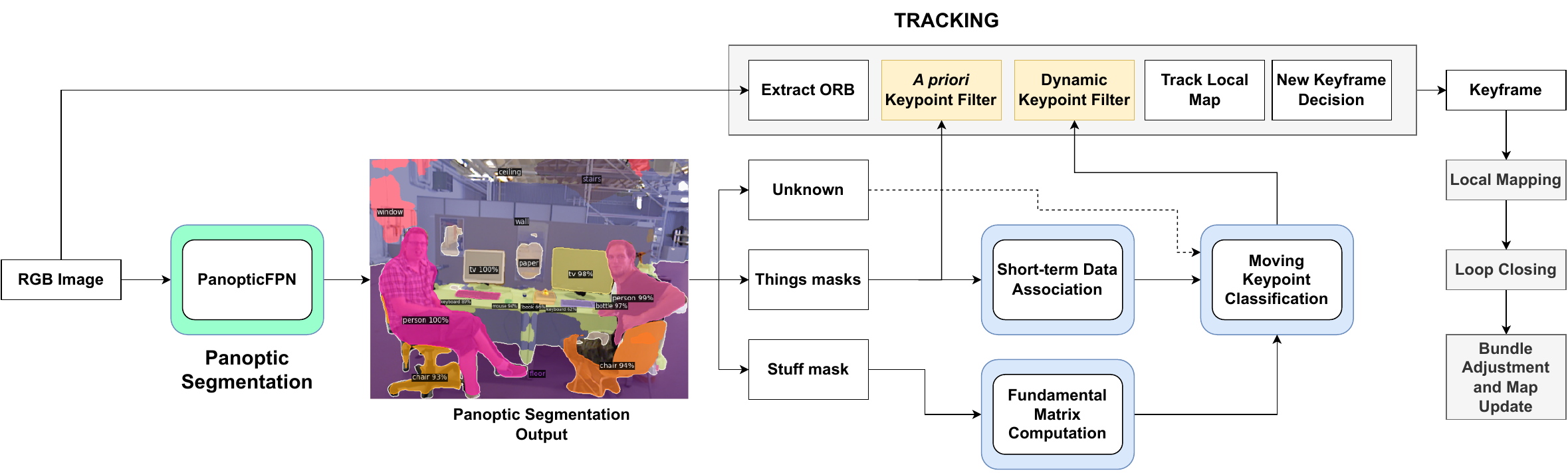}
         \caption{Framework of Panoptic-SLAM. The processes highlighted in yellow are the additional modules included in ORB-SLAM3 to allow dynamic keypoint filtering. The processes highlighted in blue describe the dynamic keypoint classification, which also run in the tracking thread. To improve computational efficiency, panoptic segmentation (in green) runs in a separate thread}
 	\label{fig:panoptic_framework}
\end{figure*}

Figure \ref{fig:panoptic_framework} shows a diagram of Panoptic-SLAM. The SLAM system is based on ORB-SLAM3 \cite{orbslam3}, and it is composed of four threads that run in parallel: panoptic segmentation, tracking, local mapping and loop closing. First, the image frames are processed in both tracking and panoptic segmentation threads. ORB features are extracted in the tracking thread, and the image is segmented into \textit{Things}, \textit{Stuff} and unknown objects in the panoptic segmentation thread. The \textit{Things} objects are sent to a short-term data association algorithm to determine if they are new or were present in the last frame. The features associated with the background (\textit{Stuff} objects) are matched with those of the last frame and used to compute a fundamental matrix. Using this fundamental matrix, the keypoints associated to known and unknown objects are classified as dynamic or static. Only static features are used for mapping and loop closing.

\subsection{Panoptic Segmentation}

In panoptic segmentation, pixels are classified either as \textit{Things} or \textit{Stuff}. \textit{Things} are countable objects with well-defined boundaries and are potentially movable such as people, animals, or vehicles. \textit{Stuff}, on the other hand, refers to uncountable amorphous regions of the image, mostly  ``unmovable" such as sky, floor, or walls \cite{coco-stuff}. The instance segmentation identifies individual objects in the \textit{Things} category, while the semantic segmentation labels all pixels in the image with their corresponding category. With high accuracy and efficiency, this segmentation method surpasses previous box-based or box-free models.

Also, some parts of the image can result in being unknown, i.e. regions that the panoptic model could not assign any label to. The unknown masks can happen due to motion blur or unlabeled objects in the scene. There can also be wrong detections, i.e., objects with wrong classification. This can happen either if an object in the scene is labeled, but there is another class with similar characteristics (e.g. tv and monitor), or if an unlabeled object is similar to a labeled class.

Similar to \cite{PVO} and \cite{fusingpanoptic}, we have opted for PanopticFPN \cite{kirillov2019panoptic} as the panoptic segmentation model, although any panoptic segmentation model can be employed in the Panoptic-SLAM framework. PanoticFPN is trained with the COCO dataset \cite{coco}, which can segment up to 80 \textit{Thing} different labels and 91 \textit{Stuff} different labels. Figure \ref{fig:2} shows an example of an output of panoptic segmentation.

\subsection{Dynamic Keypoint Filtering}

The proposed method for dynamic keypoint detection and filtering is divided into four processes: \textit{a priori} keypoint filtering, fundamental matrix computation, short-term data association, and moving keypoint classification of \textit{Things} and Unknown.

\textit{Keypoint filtering}: First, the panoptic model generates all masks predictions. People keypoints are filtered \textit{a priori}, as humans can be considered as highly dynamic, and just in rare situations they remain completely still for a long period. The keypoints belonging to the \textit{Stuff} mask are used for computing the fundamental matrix with RANSAC. 

\textit{Fundamental matrix computation}: In order to calculate the fundamental matrix, corresponding features must be identified in two consecutive frames. Feature matching involves finding pairs of points in the two frames that correspond to the same 3D point in the scene. Without an accurate feature matching, the fundamental matrix cannot be accurately calculated, which can result in errors in subsequent tasks. The feature matching algorithm matches the ORB descriptors of the keypoints in both the reference and the query images using a nearest neighbor approach. Specifically, for each keypoint in the reference image, the algorithm searches for the closest keypoint in the query image based on the similarity of their ORB descriptors.

The fundamental matrix is used to map feature points from the previous frame to their corresponding search domain in the current frame, namely the epipolar line. Assuming that the matched points in the current and last frames are $p_1$ and $p_2$, respectively, their homogeneous coordinate form can be represented as $P1$ and $P2$.
\begin{equation}
    P1_j = [u_1,v_1,1], P2_j =[u_2,v_2,1] 
\end{equation}
where $u$ and $v$ represent the pixel coordinates in the image frame, and $j$ symbolizes the keypoint category (\textit{Thing} or Unknown). Using these values, we can compute the epipolar line, denoted as $L$:
\begin{equation}
L
    \begin{bmatrix}
X\\
Y\\
Z
\end{bmatrix} = F P1_j
\end{equation}

Given that $X$, $Y$, $Z$ represent line vectors and $F$ represents the fundamental matrix, the distance between a matched point and its corresponding epipolar line can be determined as 
\begin{equation}
    D_j = \frac{P2_j^T F P1_j}{\sqrt{\vert\vert X \vert\vert ^2 + \vert\vert Y \vert\vert ^2}}
\end{equation}

\noindent where $D_j$ is the distance between a matched point and its corresponding epipolar line. The threshold value for $D_j$ was configured to 0.1, so the system could be sensitive to the movement of keypoints.

The idea is to have the fundamental matrix calculated with static keypoints and use it with \textit{Things} and unknown feature matches in order to determine whether the distance from a matched point to its corresponding epipolar line is less than a certain threshold. If the distance is lower than the threshold, the matched keypoint is considered to be static.   

\textit{Short-term data association}: The moving keypoint classification of \textit{Things} uses a short-term data association algorithm to address the issue of multiple objects with the same label in a frame. The short-term data association evaluates, using the intersection over union (IoU) metric, if a new Thing mask detected in the current frame correspond to a Thing mask of the last frame. Every instance mask is predicted with its respective bounding box. For every new pair of frames at time $k$ and $k-1$, the algorithm checks if any two bounding boxes from the same label overlap. Once an overlap is detected, the algorithm will extract the object contours $C^{k}$ and $C^{k-1}$ to compute the $IoU = |C^{k} \cap C^{k-1}| / |C^{k} \cup C^{k-1}|$ and determine if both contours have an association. To filter the keypoints of objects and determine which ones are in motion, it is essential to differentiate between multiple objects that share the same label. If an object in the current frame is not matched with any object in the previous frame, it is considered a new one and the features associated with its mask are filtered.

\textit{Moving keypoint classification}: After the data association, the current and last frame keypoints belonging to \textit{Things} from the same label and same tracking ID are matched. $D_{thing}$ is calculated using the matched features and the fundamental matrix to determine which points are dynamic and, consequently, filtered.  

\begin{figure}[h!]
  \subfigure[  \hspace{0.05cm}    Unknown mask ]{%
    \includegraphics[width=.481\linewidth]
    {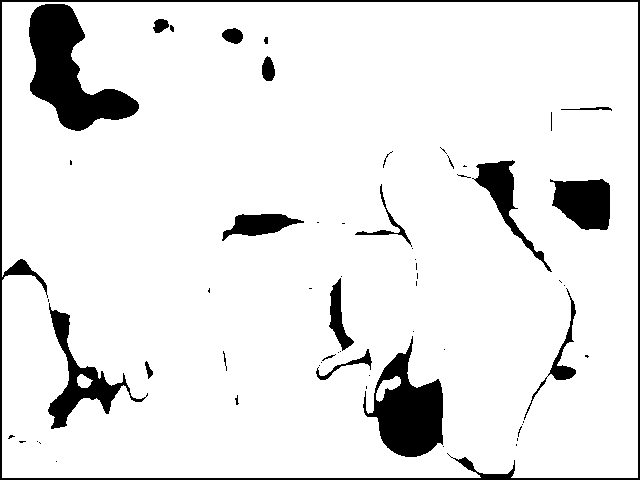}
    \label{fig:unknown} 
  } 
  \subfigure[ \hspace{0.05cm}   Keypoint filtering]{%
    \includegraphics[width=.481\linewidth]
    {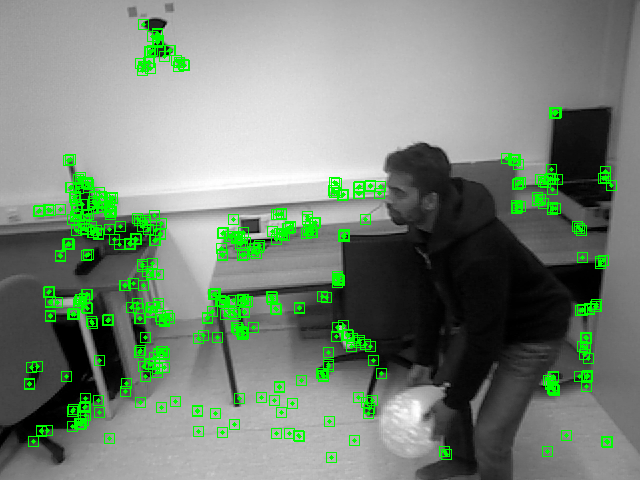}
    \label{fig:undetected_features} 
  } 

  \caption{Example of an unknown moving object being filtered in a sequence of the Bonn RGB-D dynamic dataset} 
\end{figure}

Features belonging to unknown pixels are located by adding all known masks into a single image and analyzing the black areas in it, as shown in Fig. \ref{fig:unknown}. The process is similar to the moving keypoint classification of \textit{Things}. After matched keypoints from current and last frame are found in the unknown mask, these points are checked using the fundamental matrix to calculate the epipolar distance $D_{unk}$ and to determine if they are static or dynamic. The unmatched points are filtered.

\section{RESULTS}

Sequences from two different benchmark datasets were used to evaluate the system and to compare it to state-of-the-art methods. Also, the system was tested in a real environment, measuring the ground-truth. The experimental results were also compared with other methods from the literature. All tests were performed five times and the median values were selected.

\subsection{TUM RGB-D dataset}

The TUM RGB-D dataset \cite{benchmark} is used to evaluate the robustness of the system in dynamic environments. It includes both RGB and depth image sequences captured by a Microsoft Kinect camera, along with corresponding ground truth trajectories. The data was recorded at a resolution of 640$\times$480 and a frequency of 30~Hz. Four sequences were chosen for the evaluation: fr3\_w\_static, fr3\_w\_xyz, fr3\_w\_rpy and fr3\_w\_halfsphere. These sequences depict two people walking through a room, moving behind a desk, passing in front of the camera, and sitting on chairs. The main difference between each sequence is the camera motion. In the xyz sequence, the camera is moved along the three axis while maintaining a fixed orientation. In the rpy sequence the camera is rotated around roll, pitch, and yaw axis, maintaining a fixed position. In the halfsphere sequence, the camera follows a trajectory along a half-sphere. In the static sequence, the camera remains still. The global consistency of the estimated trajectory is analyzed using the Absolute Trajectory Error (ATE) \cite{benchmark}. This metric compares the absolute distances between the translational components of ground truth and estimated trajectories.

\begin{figure}[h]
  \subfigure[  \hspace{0.05cm}    fr3\_w\_rpy sequence ]{%
    \includegraphics[width=.481\linewidth]{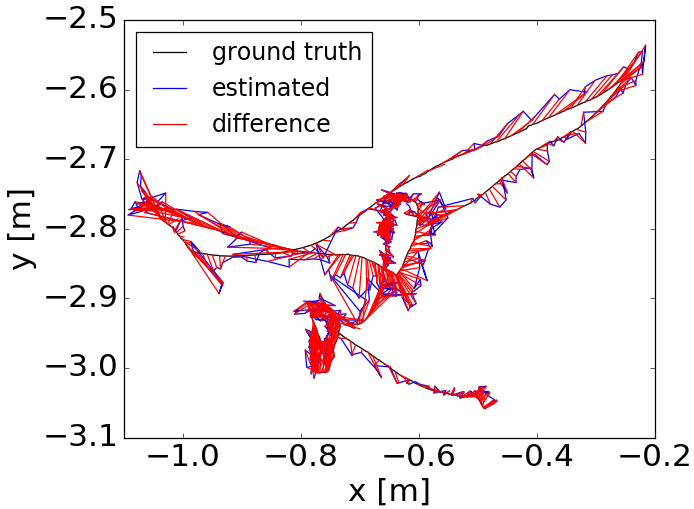}
    \label{fig:plot_tum_rpy} 
  } 
  \subfigure[ \hspace{0.05cm}   fr3\_w\_halfsphere sequence]{%
    \includegraphics[width=.481\linewidth]{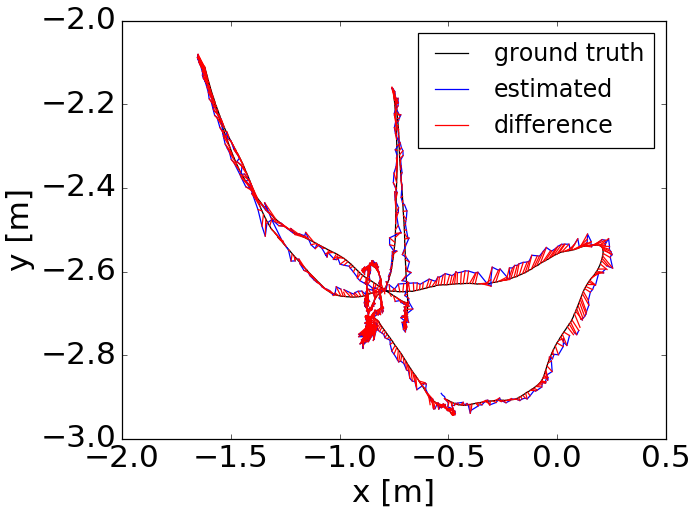}
    \label{fig:plot_tum_half} 
  } 

  \caption{Comparison between the ground-truth and the trajectory estimated by Panoptic-SLAM in two dynamic sequences of the TUM RGB-D dataset} 
\end{figure}

Figures \ref{fig:plot_tum_rpy} and \ref{fig:plot_tum_half} show the comparison between the ground-truth and estimated trajectories of Panoptic-SLAM, respectively, in the fr3\_w\_rpy and fr3\_w\_halfsphere sequences. Despite being very challenging scenes, with fast and complex camera motions, especially in the halfsphere sequence, the Panoptic-SLAM system was able to perform an accurate estimation. Table \ref{tab:ate_comparison_tum} shows the ATE comparison between Panoptic-SLAM and several methods from the literature. The best results are highlighted in bold. Based on the comparison results, Panoptic-SLAM achieved a similar accuracy of DynaSLAM. This is likely due to the fact that DynaSLAM filters people in advance, whereas the TUM fr3\_walking dataset primarily has people moving. Ji et al.~\cite{ji2021} has an inferior accuracy than ours, despite also being robust to unknown labels in the environments. Our method also outperforms Zhu et al. \cite{fusingpanoptic}, that also uses panoptic segmentation.

\begin{table}[h]

\caption{Comparison of the RMSE of ATE [m] of Panoptic-SLAM against 10 state-of-the-art methods using the TUM RGB-D dataset}

\label{tab:ate_comparison_tum}
\begin{center}
\setlength{\tabcolsep}{3.8pt}
\renewcommand{\arraystretch}{1.0}
\begin{tabular}{|c|c|c|c|c|}
\hline
Sequence  & \textit{fr3\_w\_static} & \textit{fr3\_w\_xyz} & \textit{fr3\_w\_rpy}& \textit{fr3\_w\_half} \\
\hline
Panoptic-SLAM  & 0.009 & \textbf{0.014} & \textbf{0.032}  &  \textbf{0.025} \\
\hline
ORB-SLAM3 \cite{orbslam3}   & 0.038  & 0.819 & 0.957  &  0.315 \\
\hline
PVO \cite{PVO} & 0.007 & 0.018 & 0.056 & 0.221 \\
\hline
ReFusion \cite{re-fusion}    &  0.017 & 0.099 & --- & 0.104 \\
\hline
DynaSLAM \cite{dynaslam}   &  \textbf{0.006} & 0.015 & 0.035 & \textbf{0.025} \\
\hline
DS-SLAM \cite{dsslam}  & 0.008  & 0.024 & 0.444 &  0.030 \\
\hline
SaD-SLAM \cite{sadslam}   & 0.017 & 0.017  & \textbf{0.032}  &  0.026  \\
\hline
DOTMask \cite{dotmask}    & 0.008 & 0.021  & 0.053  &  0.040 \\
\hline
Ji et al. \cite{ji2021}  & 0.011 & 0.020  & 0.037  & 0.029\\
\hline
Zhu et al. \cite{fusingpanoptic}  & 0.013& 0.018   & 0.039  & 0.030\\
\hline
DeFlowSLAM \cite{DeFlowSLAM}  & 0.007 & 0.018   & 0.057  & 0.042\\
\hline
\end{tabular}
\end{center}
\end{table}

\subsection{Bonn RGB-D dynamic dataset}

The Bonn RGB-D dynamic dataset \cite{re-fusion} is used to evaluate the robustness against moving objects, people and non-labeled moving objects. It also uses the same evaluation metrics from the TUM RGB-D dataset. Six sequences, filmed inside an indoor environment, were chosen from the dataset for evaluation: Balloon, Balloon2, non-obstructing box (non-obst box) 1 and 2, placing non-obstructing box (placing\_no\_box) 1 and 2. The Balloon sequences show a person walking and interacting with a floating balloon. In the non-obst box, a person moves a cardboard box from one place to another. In the placing\_no\_box sequences, a person appears and places a cardboard box on the floor. It is important to state that both balloon and cardboard box classes are not present in the COCO dataset \cite{coco} and,  consequently, cannot be detected by our segmentation model.

\begin{figure}[h!]
  \subfigure{%
        \includegraphics[width=.481\linewidth]{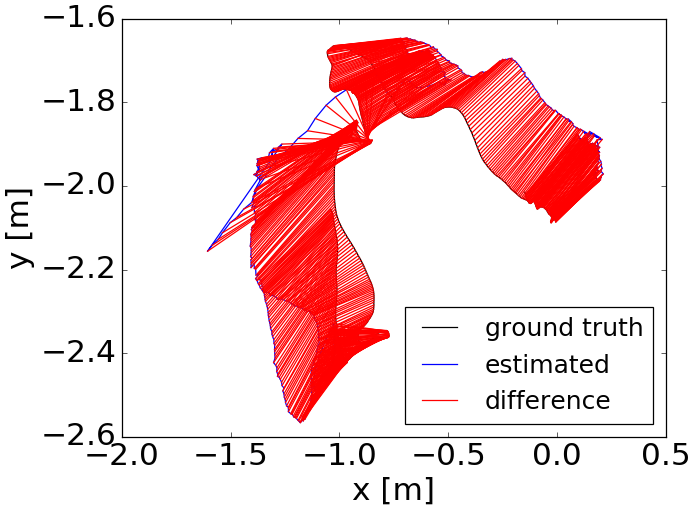}
  } 
  \subfigure{%
    \includegraphics[width=.481\linewidth]{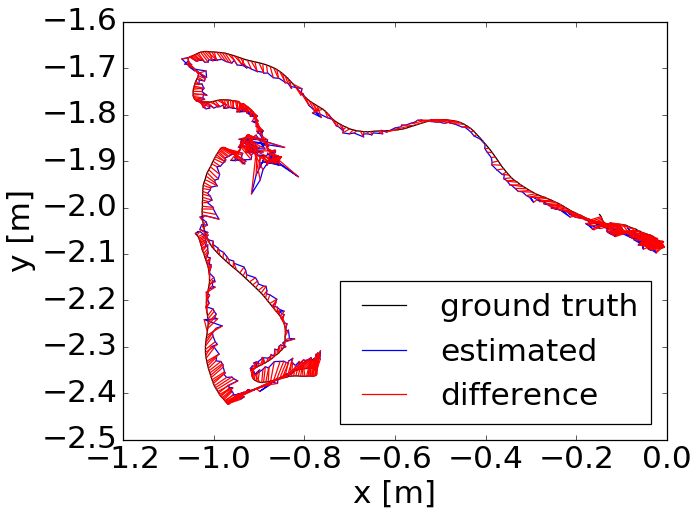}
  } 

  \caption{Comparison between the ground-truth and the trajectory estimated by ORB-SLAM3 (left) and Panoptic-SLAM (right) in the Non-obstructing box sequence of the Bonn RGB-D dynamic dataset} 
  \label{fig:plot_bonn_panoptic}
\end{figure}

Figure~\ref{fig:plot_bonn_panoptic} shows the comparison between the ground-truth and estimated trajectories of ORB-SLAM3 and Panoptic-SLAM, respectively, in the moving non-obstructing box sequence. The trajectory estimated by ORB-SLAM3 completely deviates from the ground-truth, while Panoptic-SLAM is able to perform a correct estimation. Table \ref{tab:ate_comparison_bonn} shows the RMSE of the ATE comparison between Panoptic-SLAM and ORB-SLAM3, DynaSLAM and ReFusion. The results of ReFusion and DynaSLAM were obtained in \cite{re-fusion}. Our system outperformed ORB-SLAM3 in every sequence. Our system also outperformed the other systems in every sequence, except for the last one, where it achieved similar results to DynaSLAM, with a difference of 1 mm. Despite the fact that DynaSLAM achieved similar results to ours in the TUM RGB-D dataset, the same does not occur in the Bonn RGB-D dynamic dataset due to the presence of unknown moving objects. This is evident in the results of the non-obst box and placing-no-box sequences, where the error from DynaSLAM is 8.6 and 13 times larger than ours, respectively.

\begin{table}[h!]
\caption{Comparison of the RMSE of ATE [m] of Panoptic-SLAM against ORB-SLAM3, DynaSLAM, and ReFusion using the Bonn RGB-D dynamic dataset}
\label{tab:ate_comparison_bonn}
\begin{center}
\setlength{\tabcolsep}{1.6pt}
\renewcommand{\arraystretch}{1.0}
\begin{tabular}{|c|c|c|c|c|c|c|}
\hline
Sequence & Panoptic-SLAM & ORB-SLAM3 & DynaSLAM  & ReFusion \\
\hline
Non-obst box   &  \textbf{0.027} & 0.347 & 0.232 & 0.071  \\
\hline
Non-obst box2 & \textbf{0.033} & 0.043  & 0.039  & 0.179  \\
\hline
Balloon & \textbf{0.029}  & 0.092  & 0.030 &  0.175 \\
\hline
Balloon2 & \textbf{0.027}  & 0.215  & 0.029 & 0.254  \\
\hline
placing\_no\_box & \textbf{0.044}  & 0.842 & 0.575 &  0.106  \\
\hline
placing\_no\_box2 & 0.022  & 0.023  & \textbf{0.021} &  0.141  \\
\hline

\end{tabular}
\end{center}
\end{table}

To show the importance of each step of our methodology, experiments were made using three different configurations of the system: using only the people filter, the people filter together with the known moving object filter, and the people filter together with the unknown moving object filter. Two challenging sequences of the Bonn RGB-D dynamic dataset that contain unknown moving objects were chosen for the evaluation: non-obst box and placing-non-box. Table \ref{tab:ate_configurations} shows the RMSE of the ATE of the different configurations, compared to the combined system. Due to the presence of both known and unknown moving objects in these scenes, every configuration fails except for the combined system. Using only the people filter with the unknown object filter, the results also drift because the cardboard box gets mislabeled in some frames, causing moving keypoints to not be filtered.

\begin{table}[h!]
\caption{Evaluation of the ATE on the Bonn RGB-D dynamic dataset using Panoptic-SLAM with different configurations~[m]}
\label{tab:ate_configurations}
\begin{center}
\renewcommand{\arraystretch}{1.0}
\begin{tabular}{|c|c|c|}
\hline
Sequence  & non-obst box & placing\_no\_box  \\
\hline
People filter   & 0.481  & 0.707  \\
\hline
People + Moving object filter   & 0.029  &  0.765 \\
\hline
People + Unknown object filter   & 0.302  &  0.721 \\
\hline
Panoptic-SLAM  & \textbf{0.027} & \textbf{0.044} \\
\hline
\end{tabular}
\end{center}
\end{table}

\subsection{Experiments}

\begin{figure}[h!]
  \subfigure{%
    \includegraphics[width=.467\linewidth]{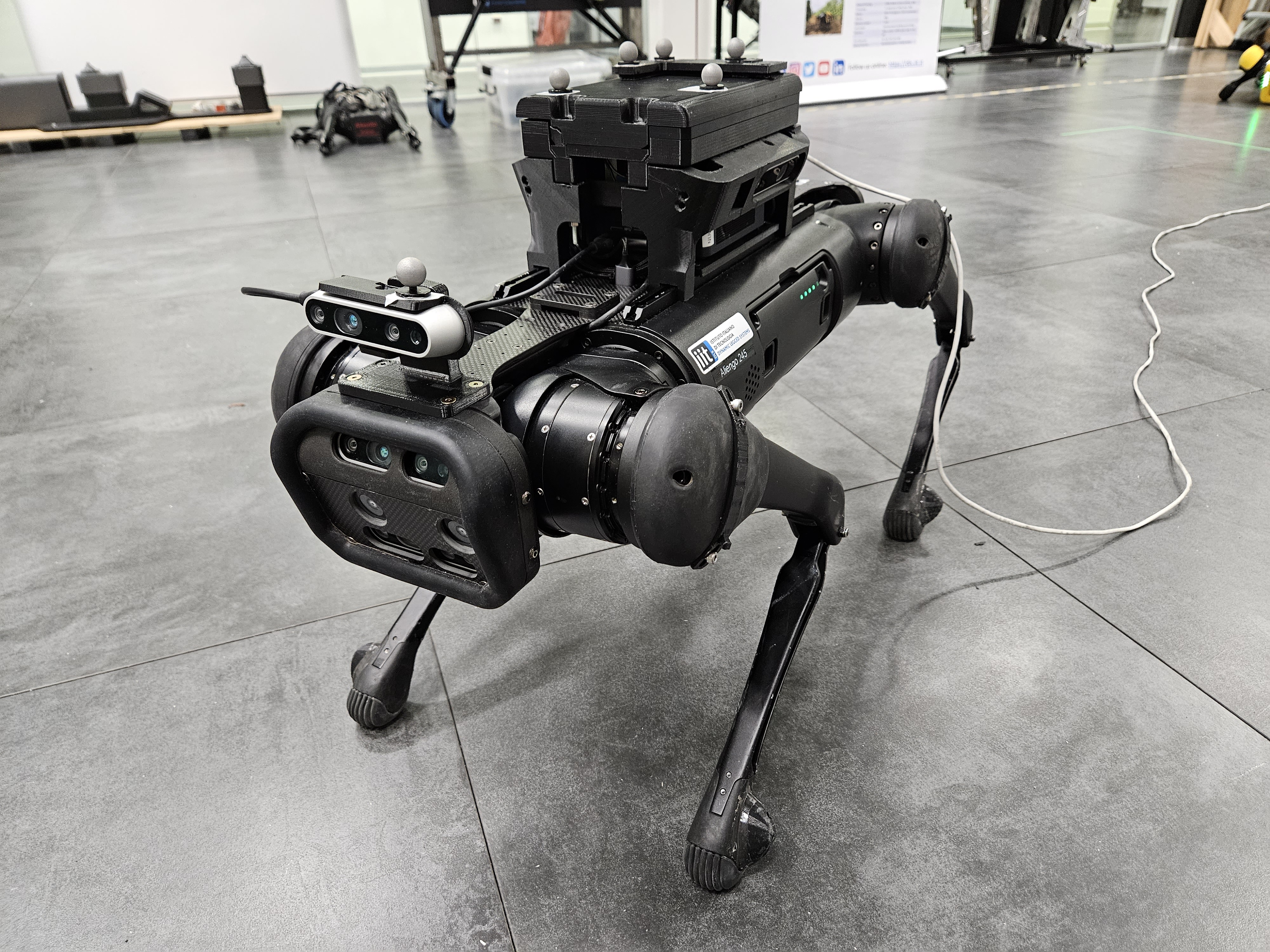}
    \label{fig:aliengo} 
  } 
  \subfigure{%
    \includegraphics[width=.489\linewidth]{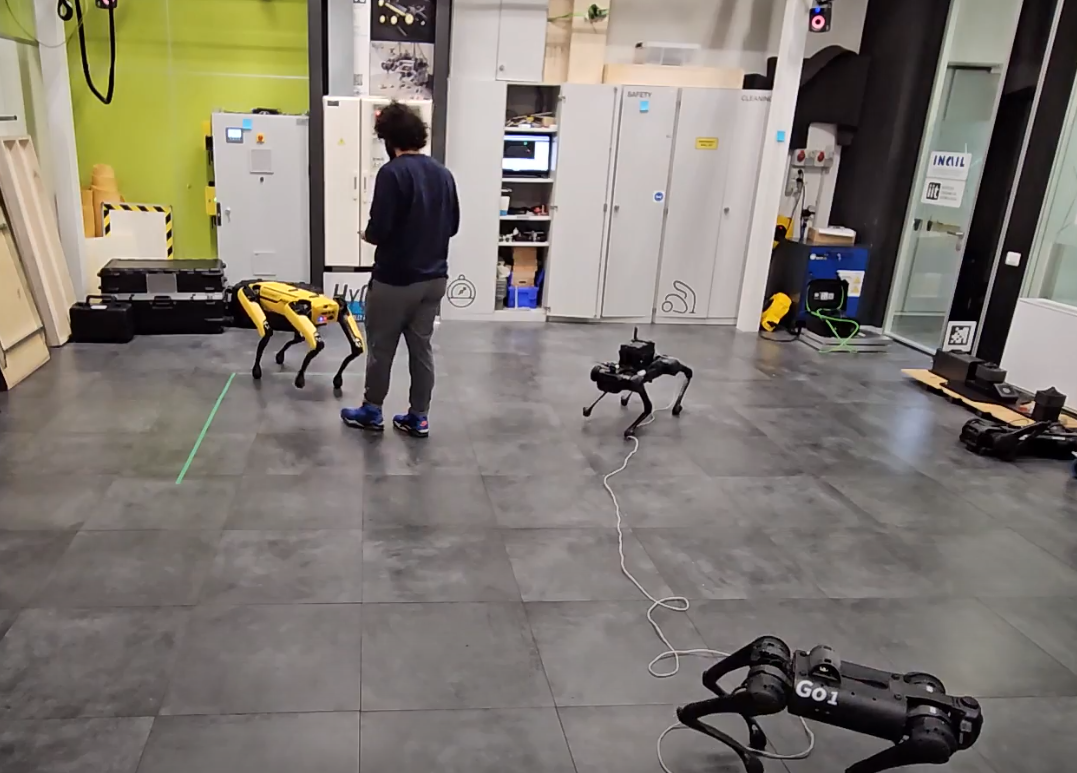}
    \label{fig:experiment_example_external}
  } 
  \caption{Images of the experiments: AlienGo quadruped robot with RGB-D camera (left). A scene of the robots\_people sequence from an external point of view (right)} 
  \label{fig:experiments}
\end{figure}

Experiments were made to show the applicability of the proposed methodology in real-world indoor scenarios. The sequences were designed to represent real situations in a laboratory or a factory, with people and other robots moving, while the robot performing SLAM is also moving in the scene. Data was recorded in an Intel NUC computer attached to a Unitree AlienGo quadruped robot equipped with a RealSense D435 camera, as shown in Fig. \ref{fig:experiments}~(left). A
Vicon motion capture system was used to obtain the ground truth.

Three different sequences were recorded. In the first one, the robot moved in a static scenario, to serve as a baseline. In the second sequence, named moving\_robots, the robot walked forward, while other robots walked nearby, namely a Spot from Boston Dynamics, and a Go1 from Unitree. Finally, in the third sequence, robots\_people, the robot walked in a closed path, with not only robots walking near it, but also people, as shown by Fig. \ref{fig:experiments}~(right).

\begin{figure}[h!]
  \subfigure[  \hspace{0.05cm} Panoptic segmentation (RP)]{%
    \includegraphics[width=.485\linewidth]{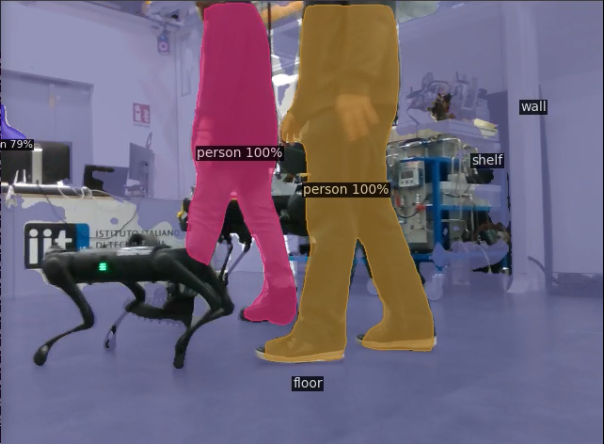}
    \label{fig:panoptic_seg_exp1} 
  } 
  \subfigure[ \hspace{0.05cm}   Keypoint filtering (RP)]{%
    \includegraphics[width=.476\linewidth]{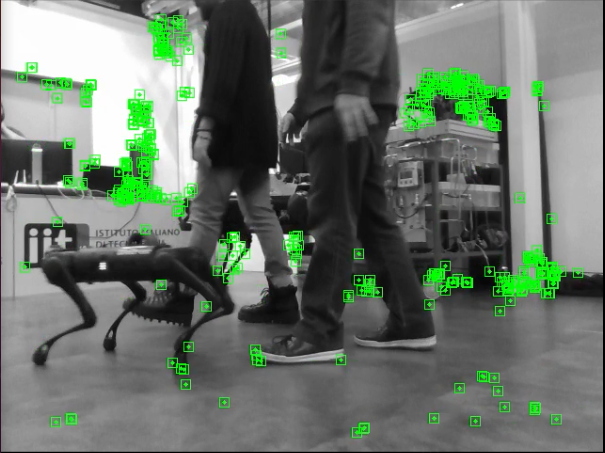}
    \label{fig:panoptic_filter_exp1}
  } 

  \subfigure[  \hspace{0.05cm} Panoptic segmentation (MR)]{%
    \includegraphics[width=.483\linewidth]{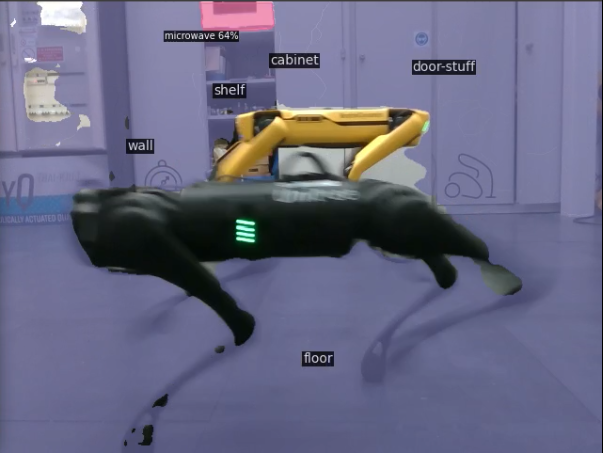}
    \label{fig:panoptic_seg_exp2} 
  } 
  \subfigure[ \hspace{0.05cm}   Keypoint filtering (MR)]{%
    \includegraphics[width=.479\linewidth]{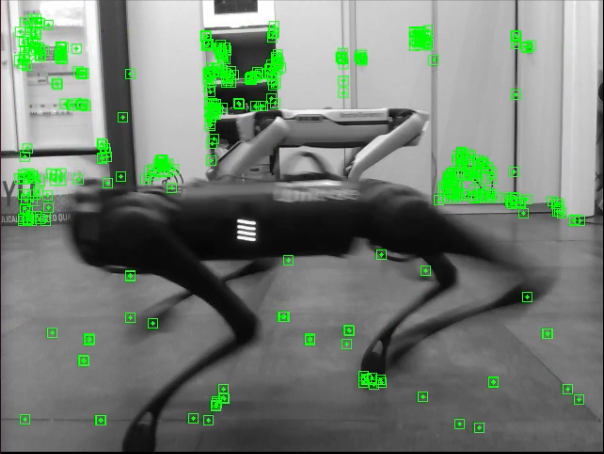}
    \label{fig:panoptic_filter_exp2}
  } 
 \caption{Output examples of the panoptic segmentation and keypoint filtering steps from Panoptic-SLAM in the sequences robots\_people (RP) and moving\_robots (MR)} 
\end{figure}

Figures \ref{fig:panoptic_seg_exp1}-\ref{fig:panoptic_filter_exp2} show the panoptic segmentation and keypoint filtering in two frames captured during the experiments. In Fig. \ref{fig:panoptic_seg_exp1} it is possible to differentiate the segmented people, the background and the quadruped robot Go1, which is unlabeled. With this information, it is possible to filter the keypoints belonging to the moving parts of the scene, as shown in Fig. \ref{fig:panoptic_filter_exp1}. The same behavior is observable in Figs.~\ref{fig:panoptic_seg_exp2} and \ref{fig:panoptic_filter_exp2}.

\begin{figure}[h!]
  \subfigure[  \hspace{0.05cm}    ORB-SLAM3 (B)]{%
    \includegraphics[width=.481\linewidth]{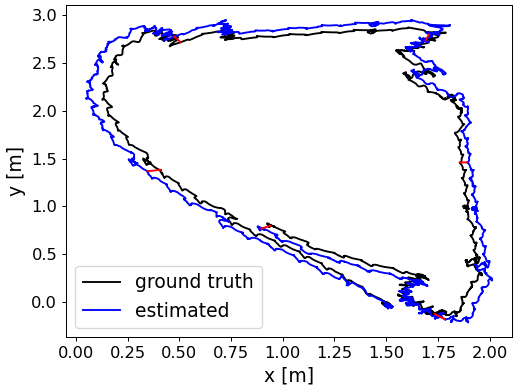}
    \label{fig:ATE_orbslam3_baseline} 
  } 
  \subfigure[ \hspace{0.05cm}   Panoptic-SLAM (B)]{%
    \includegraphics[width=.481\linewidth]{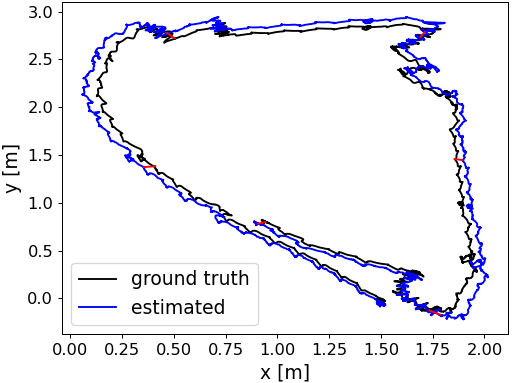}
    \label{fig:ATE_panopticslam_baseline} 
  }

  \subfigure[  \hspace{0.05cm}    ORB-SLAM3 (MR)]{%
    \includegraphics[width=.481\linewidth]{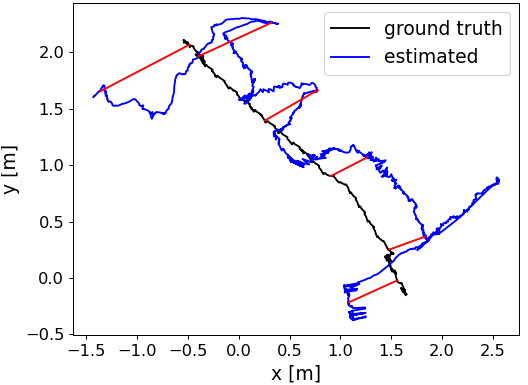}
    \label{fig:ATE_orbslam3_slamrobots2} 
  } 
  \subfigure[ \hspace{0.05cm}   Panoptic-SLAM (MR)]{%
    \includegraphics[width=.465\linewidth]{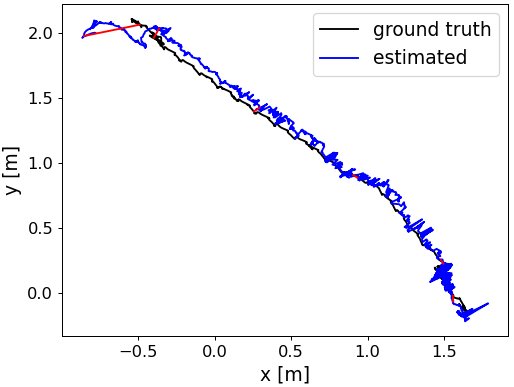}
    \label{fig:ATE_panoptic_slamrobots2} 
  }


  \subfigure[  \hspace{0.05cm}    ORB-SLAM3 (RP)]{%
    \includegraphics[width=.483\linewidth]{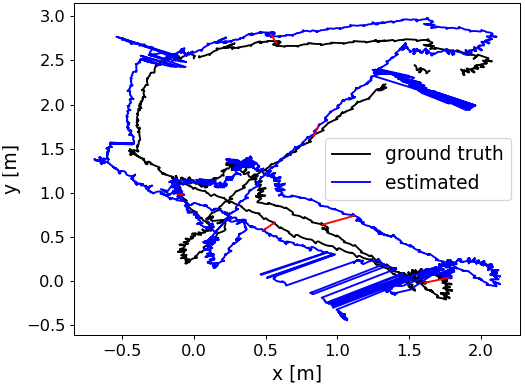}
    \label{fig:ATE_orbslam3_people} 
  } 
  \subfigure[ \hspace{0.05cm}   Panoptic-SLAM (RP)]{%
    \includegraphics[width=.471\linewidth]{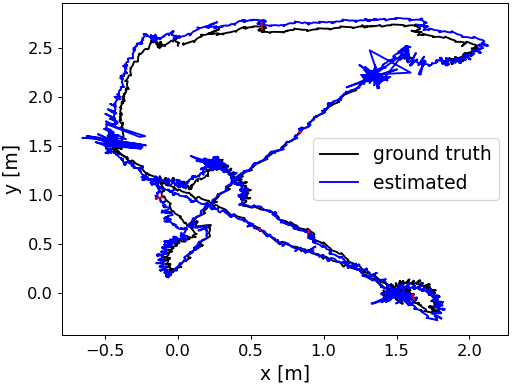}
    \label{fig:ATE_panopticslam_people} 
  } 
  \caption{Comparison between the ground-truth and the trajectory estimated by ORB-SLAM3 and Panoptic-SLAM in the sequences baseline (B), moving\_robots (MR), and robots\_people (RP)} 

  \subfigure[ \hspace{0.05cm}   moving\_robots]{%
    \includegraphics[width=.481\linewidth]{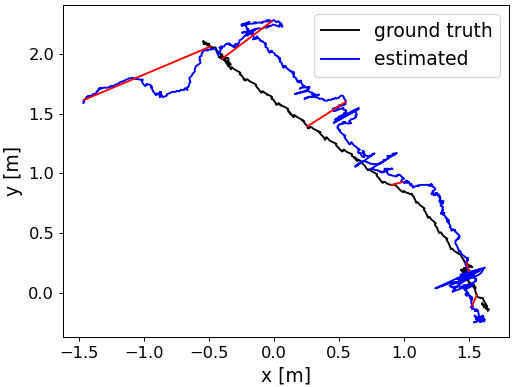}
    \label{fig:ATE_dyna_robots} 
  }   
  \subfigure[  \hspace{0.05cm}    robots\_people ]{%
    \includegraphics[width=.481\linewidth]{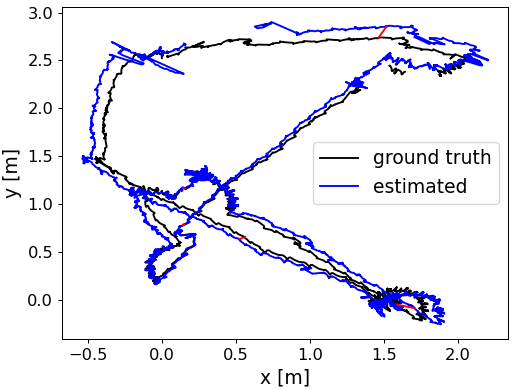}
    \label{fig:ATE_dyna_people} 
  } 
  \caption{Comparison between the ground-truth and the trajectory estimated by DynaSLAM in the moving\_robots and robots\_people sequences} 
\end{figure}

Table \ref{tab:ate_comparison_experiments} shows the RMSE of the ATE comparison between Panoptic-SLAM, ORB-SLAM3, and DynaSLAM using the data recorded in the experiments. The best results are highlighted in bold. Our system outperformed ORB-SLAM3 in every sequence. As expected, the values for the baseline sequence are closer to each other. In the other sequences, on the other hand, ORB-SLAM3 has an increased error due to its incapability of filtering dynamic content. This is also evident in Figs. \ref{fig:ATE_orbslam3_baseline} - \ref{fig:ATE_panopticslam_people}, that show the comparison between the ground-truth and estimated trajectories of ORB-SLAM3 and Panoptic-SLAM, respectively, in the experiments. Figures \ref{fig:ATE_orbslam3_slamrobots2} and \ref{fig:ATE_orbslam3_people} show how the trajectories estimated by ORB-SLAM3 deviate from the ground-truth in several parts. Panoptic-SLAM is able to maintain a low-drift estimation, as shown in Figs. \ref{fig:ATE_panoptic_slamrobots2} and \ref{fig:ATE_panopticslam_people}.

\begin{figure}[h!]

  \subfigure[ \hspace{0.05cm}   DynaSLAM (RP)]{%
    \includegraphics[width=.471\linewidth]{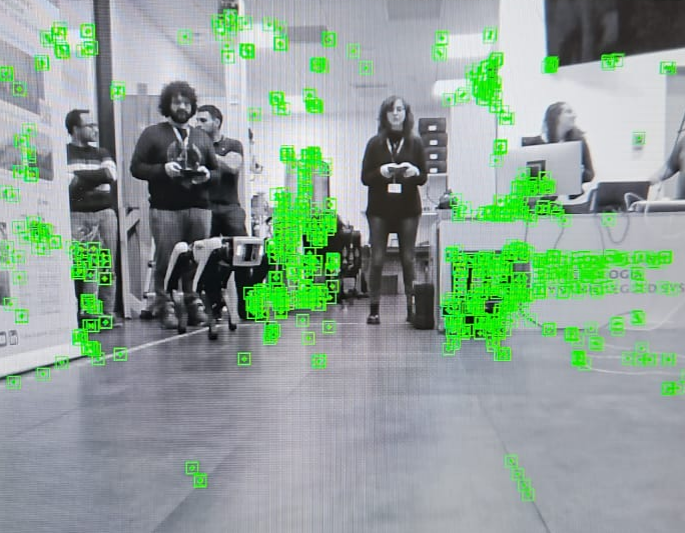}
    \label{fig:dyna_keypoints}
  } 
  \subfigure[  \hspace{0.05cm} Panoptic-SLAM (RP)]{%
    \includegraphics[width=.438\linewidth]{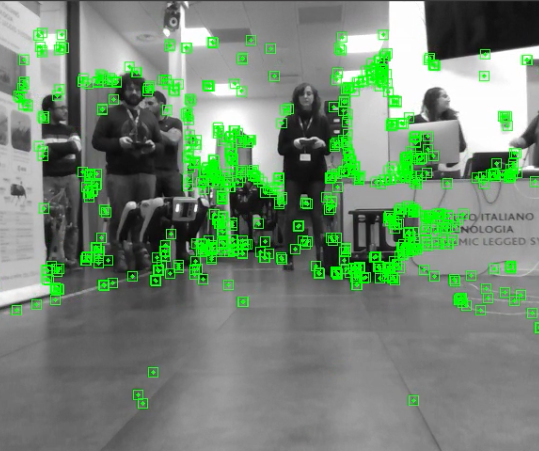}
    \label{fig:panoptic_seg_keypoint} 
  }

  \subfigure[ \hspace{0.05cm}   DynaSLAM (MR)]{%
    \includegraphics[width=.46\linewidth]{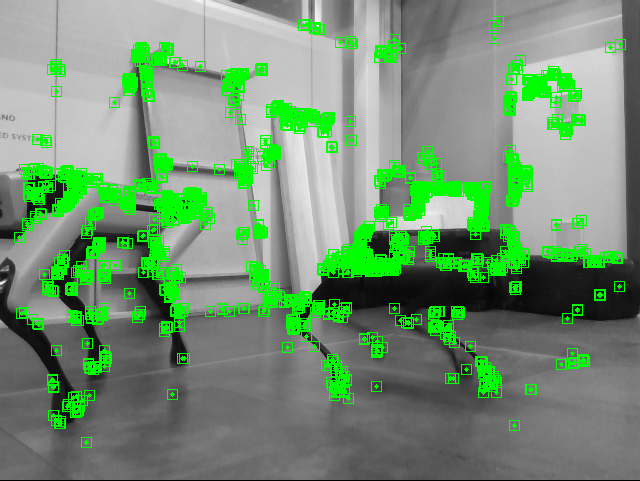}
    \label{fig:dyna_keypoints2}
  } 
  \subfigure[  \hspace{0.05cm} Panoptic-SLAM (MR)]{%
    \includegraphics[width=.46 \linewidth]{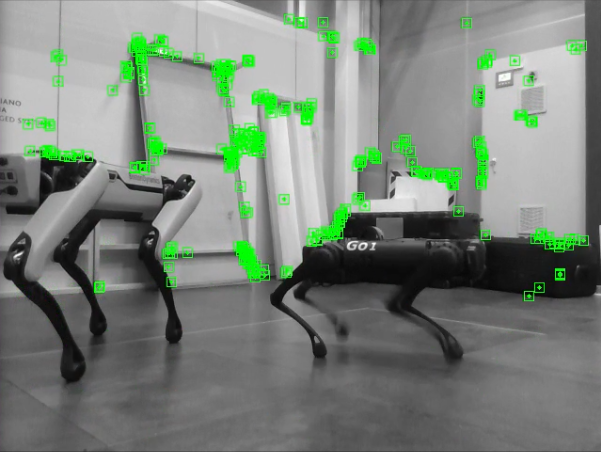}
    \label{fig:panoptic_seg_keypoint2} 
  } 
  \caption{Keypoint filtering comparison between DynaSLAM and Panoptic-SLAM in the sequences robots\_people (RP) and moving\_robots (MR)} 
\end{figure}

Figures \ref{fig:ATE_dyna_robots} and \ref{fig:ATE_dyna_people} display the results from DynaSLAM in the moving\_robots and robots\_people experiments, respectively. Due to the lack of robustness to detect unlabeled moving objects, DynaSLAM could not accurately estimate its pose in the moving\_robots sequence. In the other sequence, loop closure detection was important to lower the pose estimation drifts. Furthermore, DynaSLAM's people mask dilation helped to filter most of the dynamic content in the scene. However, it also wrongly filters static keypoints. This can be seen by comparing the keypoint filtering from DynaSLAM and Panoptic-SLAM in Figs.  \ref{fig:dyna_keypoints} and \ref{fig:panoptic_seg_keypoint}. In the moving\_robots sequence, on the other hand, DynaSLAM presented a higher error since it is not capable of filtering the dynamic keypoints due to the absence of people in the scene, as shown in Fig. \ref{fig:dyna_keypoints2}.

\begin{table}[h!]
\caption{Comparison of the RMSE of ATE [m] of Panoptic-SLAM against ORB-SLAM3 and DynaSLAM in the experiments}
\label{tab:ate_comparison_experiments}
\begin{center}
\setlength{\tabcolsep}{1.6pt}
\renewcommand{\arraystretch}{1.0}
\begin{tabular}{|c|c|c|c|c|c|c|}
\hline
Sequence & Panoptic-SLAM & ORB-SLAM3 & DynaSLAM \\
\hline
baseline  &  \textbf{0.057} & 0.062 & 0.064  \\
\hline
moving\_robots & \textbf{0.090}  & 0.495  & 0.291  \\
\hline
robots\_people & \textbf{0.055}  & 0.280  & 0.076  \\
\hline
\end{tabular}
\end{center}
\end{table}

\subsection{Run-time Analysis}

All tests were performed on a notebook with Intel i7 CPU with 16 GB of RAM and NVIDIA RTX3060 GPU with 8 GB of VRAM running Ubuntu 20.04 LTS Linux. The SLAM system is implemented in C++, and the panoptic segmentation inference is implemented in Python, using the Detectron2 framework \cite{wu2019detectron2}. 
Table \ref{tab:frame_rate_panoptic} shows the mean tracking time of the system in four sequences of the TUM RGB-D dataset, including the inference time of the panoptic segmentation. The average inference time of the panoptic segmentation was 0.2 second per frame for every sequence.

\begin{table}[h!]
\caption{Mean tracking time [s] of Panoptic-SLAM in the sequences of the TUM RGB-D dataset}
\label{tab:frame_rate_panoptic}
\begin{center}
\renewcommand{\arraystretch}{1.0}
\begin{tabular}{|c|c|}
\hline
Sequence &  Mean tracking time [s]\\
\hline
fr3\_w\_static & 0.344 \\
\hline
fr3\_w\_xyz & 0.344 \\
\hline
fr3\_w\_rpy  &  0.319 \\
\hline
fr3\_w\_half &  0.323 \\
\hline
\end{tabular}
\end{center}
\end{table}

\subsection{Limitations}

As a limitation, the computational cost of the panoptic segmentation currently does not allow Panoptic-SLAM to run in real time. Furthermore, the panoptic segmentation results can have inconsistencies due to fast camera movements and changes in illumination. However, these inconsistencies have little effect on the overall result, since the filtering is based on the movement from the matched keypoints between frames, not the object classes.

\section{CONCLUSIONS}

This paper presented Panoptic-SLAM, an open-source visual SLAM system built on ORB-SLAM3 that operates online and is robust in dynamic environments, even in the presence of unknown and unlabeled moving objects. A panoptic scene segmentation is proposed in order to classify dynamic and static keypoints. We show the effectiveness of our method by comparing it with several systems from the literature that are considered to have the highest accuracy in dynamic environments, including DynaSLAM, DS-SLAM, SaD-SLAM, PVO and FusingPanoptic in challenging dynamic sequences. The results indicate that Panoptic-SLAM not only achieves the same levels of accuracy as DynaSLAM in highly dynamic scenes, but also surpasses it by a high margin in scenarios with unknown moving objects. Furthermore, Panoptic-SLAM was tested in experiments recorded with an RGB-D camera attached to a quadruped robot. The experiments consisted of highly dynamic scenes with people moving with other quadruped robots. These robots were used to show the robustness of our system to filter non-labeled objects by the segmentation network. 

For future works, we plan to deal with moving objects that occupy a large portion of the image in order to avoid lost tracks. Also, we aim to include in the method a semantic map building that can adapt over time, further exploring the capabilities of panoptic segmentation.

\section*{\fontsize{8.5pt}{12pt}\selectfont{ACKNOWLEDGMENT}}
{
\fontsize{8.5pt}{12pt}\selectfont
The authors acknowledge financial support from PNRR MUR Project PE000013 ``Future Artificial Intelligence Research (FAIR)", funded by the European Union – NextGenerationEU.
}

\bibliographystyle{IEEEtran}
\bibliography{main}

\end{document}